\documentclass[11pt,a4paper]{article}
\usepackage[hyperref]{emnlp2020}
\usepackage{times}
\usepackage{latexsym}

\usepackage{amsmath,amsfonts,amssymb,amsthm}
\usepackage{bm}
\usepackage{comment}
\usepackage{graphicx}
\usepackage{here}
\usepackage{color}
\usepackage{url}
\usepackage{mathtools}

\usepackage{microtype}

\aclfinalcopy 

\title{Efficient Estimation of Influence of a Training Instance}

\author{Sosuke Kobayashi\textsuperscript{1,2} \quad Sho Yokoi\textsuperscript{1,3} \quad Jun Suzuki\textsuperscript{1,3} \quad Kentaro Inui\textsuperscript{1,3} \\
  Tohoku University\textsuperscript{1} \quad
  Preferred Networks, Inc.\textsuperscript{2} \quad
  RIKEN\textsuperscript{3} \\
  \texttt{sosk@preferred.jp} \\
  \texttt{\{yokoi,jun.suzuki,inui\}@ecei.tohoku.ac.jp}  }

\date{}

\begin{document}
\maketitle
\begin{abstract}
Understanding the influence of a training instance on a neural network model leads to improving interpretability.
However, it is difficult and inefficient to evaluate the influence, which shows how a model's prediction would be changed if a training instance were not used.
In this paper, we propose an efficient method for estimating the influence.
Our method is inspired by dropout, which zero-masks a sub-network and prevents the sub-network from learning each training instance.
By switching between dropout masks, we can use sub-networks that learned or did not learn each training instance and estimate its influence.
Through experiments with BERT and VGGNet on classification datasets, we demonstrate that the proposed method can capture training influences, enhance the interpretability of error predictions, and cleanse the training dataset for improving generalization.
\end{abstract}


\section{Introduction}

What is the influence of a training instance on a machine learning model?
This question has attracted the attention of the community~\cite{cook1977detection, koh17understanding,zhang2018trust,hara19dataclean}.
Evaluating the influence of a training instance leads to more interpretable models and other applications like data cleansing.

A simple evaluation is by comparing a model with another similarly trained model, whose training does not include the instance of interest.
This method, however, requires computational costs of time and storage depending on the number of instances, which indicates the extreme difficulty (Table~\ref{tab:computation_order_table}).
While computationally cheaper estimation methods have been proposed~\cite{koh17understanding,hara19dataclean}, they still have computational difficulties or restrictions of model choices.
The contribution of this work is to propose an estimation method, which (i) is computationally more efficient while (ii) useful for applications (iii) without significant sacrifice of model performance.

We propose a trick for enabling a neural network without restrictions to estimate the influence, which we refer to as \emph{turn-over dropout}.
This method is computationally efficient as it requires only running two forward computations after training a single model on the entire training dataset.
In addition to the efficiency, we demonstrated that it enabled BERT~\cite{devlin2019bert} and VGGNet~\cite{simonyan15vgg} to analyze the influences of training through various experiments, including example-based interpretation of error predictions and data cleansing to improve the accuracy on a test set with a distributional shift.


\section{Influence of a Training Instance}\label{sec:influence_formulation}

\subsection{Problem Setup}

We present preliminaries on the problem setup.
In this paper, we deal with the \emph{influence} of training with an instance on prediction with another one, which has been studied in \citet{koh17understanding}, \citet{hara19dataclean} and so on.
Let $z:=(x, y)$ be an instance and represent a pair of input $x \in X$ and its output $y \in Y$, and let $D \coloneqq \{z_i\}_{i=1}^N$ be a training dataset.
By using an optimization method with $D$, we aim to find a model $f_D\colon X \to Y$.
Denoting the loss function by $L(f, z)$, the learning problem is obtaining $\hat{f}_D = \mathrm{argmin}_{f} \mathbb{E}_{z_i \in D} L(f, z_i)$.

The \textit{influence}, $I(z_{\mathrm{target}}, z_i; D)$, is a quantitative benefit from $z_i$ to prediction of $z_{\mathrm{target}}$.
Let $f_{D\setminus\{z_i\}}$ to be a model trained on the dataset $D$ excluding $z_i$, the influence is defined as
\begin{align}
    & I(z_{\mathrm{target}}, z_i; D) \nonumber\\ & \coloneqq L(f_{D\setminus\{z_i\}}, z_{\mathrm{target}}) - L(f_{D}, z_{\mathrm{target}}) \label{eq:influence} .
\end{align}
Intuitively, the larger this value, the more strongly a training instance $z_{i}$ contributes to reduce the loss of prediction on another instance $z_{\mathrm{target}}$.
The instance of interest $z_{\mathrm{target}}$ is typically an instance in a test or validation dataset.

\begin{table}[t]
\small
\setlength\tabcolsep{3.5pt}
\begin{tabular}{llll}
\hline
Method        & Training                & Storage & Estimation \\ \hline
Re-train & $O(|D|^2)$ & $O(|\theta||D|)$   & $O(F|D|)$             \\ \hline
Hara+ & $O(|D|)$                    & $O(|\theta|T)$   & $O(F|D| + (F{+}F')TB)$       \\ \hline
Koh+          & $O(|D|)$                    & $O(|\theta|)$    & $O(F|D| + (F{+}F')rtb)$       \\ \hline
\textbf{Ours}          & \textbf{$O(|D|)$}                    & \textbf{$O(|\theta|)$}    & \textbf{$O(F|D|)$}            \\ \hline
\end{tabular}
\caption{Comparison of computational complexity for estimating the influence of all instance on another instance, with \citet{hara19dataclean} and \citet{koh17understanding}, where $|\theta|$ is the number of parameters, $F$ is a forward/backward computation, $F'$ is a double backward computation, $T$ is the training steps, $B$ is a training minibatch size, $b$ is a minibatch size for stabilizing approximation, $rt$ are the hyper-parameters; typically $rt \approx |D|$. See the references in detail.}
\label{tab:computation_order_table}
\end{table}

\subsection{Related Methods}

Computing the influence in Equation (\ref{eq:influence}) by re-training two models for each instance is computationally expensive, and several estimation methods are proposed.
\citet{koh17understanding} proposed an estimation method that assumed a strongly convex loss function and a global optimal solution\footnote{
Strictly speaking, \citet{koh17understanding} studied a similar but different value from $I$ in Equation (\ref{eq:influence}).
Briefly, the formulation in \citet{koh17understanding} considers convex models with the optimal parameters for $f_{D\setminus\{z_i\}}$ and $f_{D}$.
The definition in \citet{hara19dataclean} did not have such conditions and treated the broader problem.
We follow \citet{hara19dataclean}; therefore, the definition in Equation (\ref{eq:influence}) allows any $f_D$ and $f_{D\setminus\{z_i\}}$, as long as they have the same initial parameters and optimization procedures using the same mini-batches except for $z_i$.
}.
While the method is used even with neural models~\cite{koh17understanding,han2020explaining}, which do not satisfy the assumption, it still requires high computational cost.
\citet{hara19dataclean} proposed a method without these restrictions; however, it consumes large disk storage and computation time that depend on the number of optimization steps.
Our proposed method is much more efficient,
as shown in Table~\ref{tab:computation_order_table}.
For example, in a case where \citet{koh17understanding}'s method took 10 minutes to estimate the influences of 10,000 training instances on another instance with BERT~\cite{han2020explaining}, our method only required 35 seconds\footnote{
For the details, see Appendix~\ref{sec:appendix_kohhan_comparison}.
}.
This efficiency will expand the scope of applications of computing influence.
For example, it would enable real-time interpretation of model predictions for users of the machine learning models.

\begin{figure}[t]
  \begin{center}
  \includegraphics[width=7.7cm]{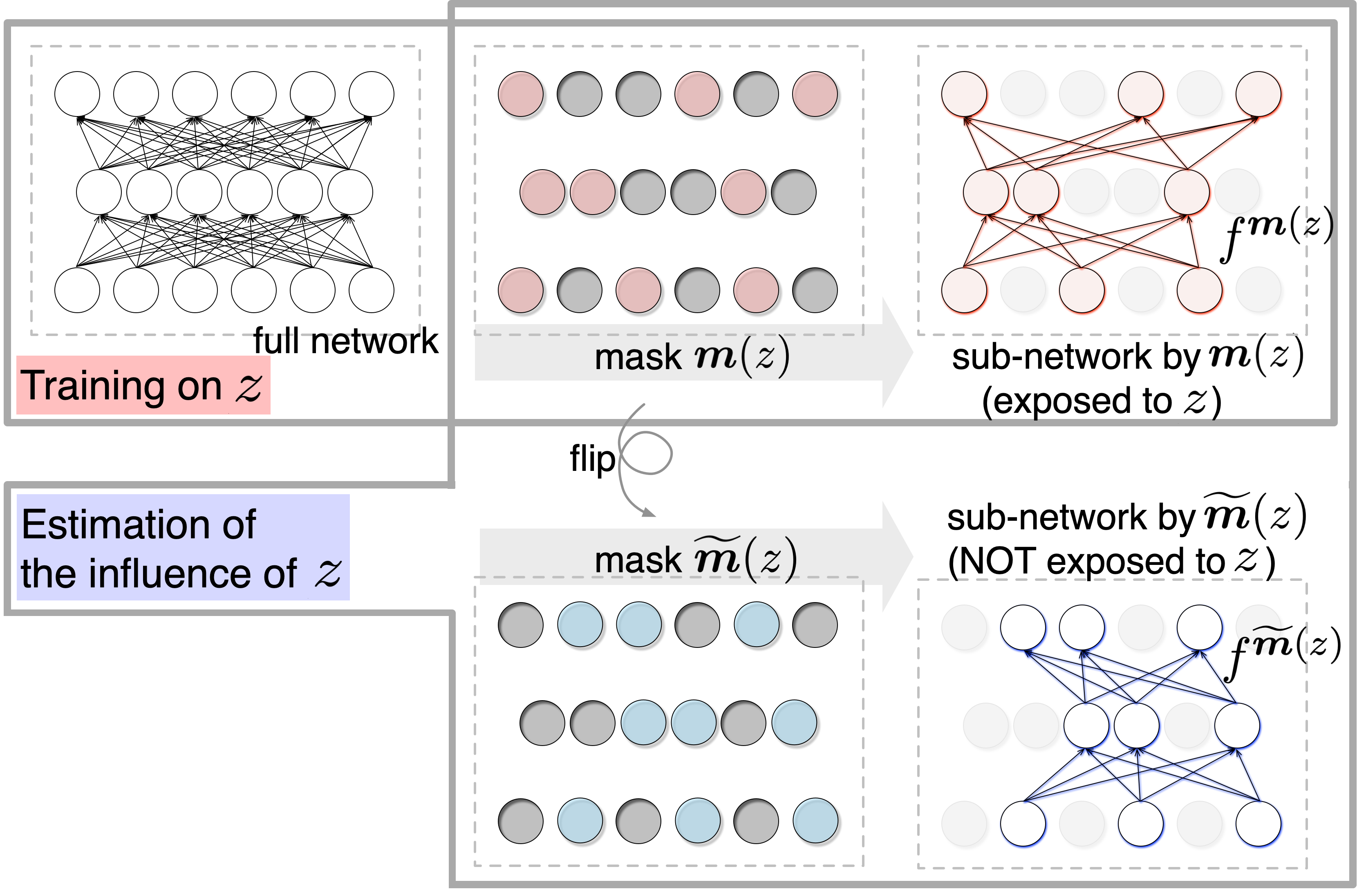}
    \caption{Dropout generates a sub-network for each training instance $z$, and updates its parameters (red; top) only. By contrast, the (blue; bottom) sub-network is not influenced by $z$. Our estimation uses the difference between the two sub-networks.}
    \label{fig:dropout_fig_more}
  \end{center}
\end{figure}


\section{Proposed Method}

\subsection{Background: Dropout}

Dropout~\cite{hinton12dropout,srivastava14dropout} is a popular regularization methods for deep neural networks.
During training, $d$-dimensional random mask vector $\boldsymbol{m}$, where $d$ refers to the number of parameters of a layer, is sampled, and a neural network model $f$ is transformed into a variant $f^{\boldsymbol{m}}$ with a parameter set multiplied with $\boldsymbol{m}$ each update\footnote{
Typically, dropout is applied to the layers of the neural network rather than its parameter matrices.
In this case, each instance in a minibatch drops different column-wise parameters of matrices at once.}.
The elements of mask $\boldsymbol{m} \in \{0, \frac{1}{p}\}^d$ are randomly sampled as follows:
$m_j := m'_j/p,\, m'_j \sim \textrm{Bernoulli}(p)$.
Parameters masked (multiplied) with $0$ are disabled in an update step like pruning.
Thus, dropout randomly selects various sub-networks $f^{\boldsymbol{m}}$ to be updated at every step.
During inference at test time, dropout is not applied.
One interpretation of dropout is that it trains numerous sub-networks and uses them as ensemble~\cite{hinton12dropout,srivastava14dropout,bachman2014,baldi2014,bulo16dropoutdistillation}.
In this work, $p=0.5$; approximately half of the parameters are zero-masked.


\subsection{Proposed Method: Turn-over Dropout}

In the standard dropout method, dropout masks are sampled independently at every update.
In our proposed method, however, we use \emph{instance-specific dropout masks} $\boldsymbol{m}(z)$, which are also random vectors but \emph{deterministically} generated and tied with each instance $z$.
Thus, when the network is trained with an instance $z$, only a deterministic subset of its parameters is updated, as shown in Figure~\ref{fig:dropout_fig_more}.
In other words, the sub-network $f^{\boldsymbol{m}(z)}$ is updated; however the corresponding counterpart of the network $f^{\widetilde{\boldsymbol{m}}(z)}$ is not at all affected by $z$, where $\widetilde{\boldsymbol{m}}(z)$ is the \emph{flipped} mask of $\boldsymbol{m}(z)$, i.e., $\widetilde{\boldsymbol{m}}(z) \coloneqq \frac{1}{p} - \boldsymbol{m}(z)$.
Both sub-networks, $f^{\boldsymbol{m}(z)}$ and $f^{\widetilde{\boldsymbol{m}}(z)}$, can be used by applying the individual masks to $f$.
These sub-networks are analogously comprehended as two different networks trained on a dataset with or without an instance, respectively, $f_{D}$ and $f_{D\setminus\{z_i\}}$\footnote{
In this paper, we associate  $f^{\boldsymbol{m}(z)}$ and $f^{\widetilde{\boldsymbol{m}}(z)}$ with $f_{D}$ and $f_{D\setminus\{z_i\}}$, respectively.
However, while $f_{D}$ does not focus on any instance in $D$ so much, its substitute $f^{\boldsymbol{m}(z)}$ may be a little biased to some characteristic of $z$.
For ignoring bias, we can use $f_{D}$ itself (i.e., full network) instead of $f^{\boldsymbol{m}(z)}$, while the representation powers of $f_{D}$ and $f_{D\setminus\{z_i\}}$ are different.
We tested the alternative but did not find large improvements.
Further exploration is an interesting future work.
}.
From this analogy, the influence of a training instance can be evaluated by considering these two sub-networks.
The influence $I(z_{\mathrm{target}}, z_i; D) = L(f_{D\setminus\{z_i\}}, z_{\mathrm{target}}) - L(f_{D}, z_{\mathrm{target}})$ is estimated as
\begin{align}
    & \hat{I}(z_{\mathrm{target}}, z_i; D) \notag \\
    & \coloneqq L(f_{D}^{\widetilde{\boldsymbol{m}}(z_i)}, z_{\mathrm{target}}) - L(f_{D}^{\boldsymbol{m}(z_i)}, z_{\mathrm{target}}) ,
\end{align}
which corresponds to the gain when using $f_{D}^{\boldsymbol{m}(z_i)}$, instead of $f_{D}^{\widetilde{\boldsymbol{m}}(z_i)}$ for a prediction on $z_{\mathrm{target}}$.
We call this estimation method \emph{turn-over dropout}.
Its summarized advantages are as follows:
{\setlength{\leftmargini}{14pt}
\begin{itemize}\vspace{-0.5\baselineskip}
\setlength{\itemsep}{2pt}
\setlength{\parskip}{0pt}
\setlength{\itemindent}{0pt}
\item \textbf{Lower computation time}: The method only requires running forward procedure two times.
\item \textbf{No snapshot or re-training}: A single model can be used for all training instances.
\item \textbf{Easy to implement}: The model modification and estimation procedure are very simple.
\end{itemize}
}


\subsection{Memory-efficient Instance-specific Masks}\label{sec:mask_construction}

One may think that using instance-specific masks require a large space, depending on the dataset size and the number of parameters to be masked.
However, this cost is drastically reduced to a constant $O(1)$, using a trick.
As the masks are not updated, we do not have to save them directly.
Instead, we can deterministically generate the random masks with a fixed random seed number anytime.
Thus, models can avoid storing masks and generate masks when using them.
We call this trick as \emph{volatile mask generation}\footnote{
The volatile mask generation method solved storage and memory issues in our experiments.
However, the memory issue could occur even with the method, depending on implementations.
For such a particular case and another solution for it, see Appendix~\ref{sec:appendix_hashmask} in detail.}.
Simultaneously, \citet{wortsman2020supermasks} uses this trick for generating task-specific sub-networks.


\section{Experiments}\label{sec:experiments}

The computational efficiency of our method is discussed in Section~\ref{sec:influence_formulation}.
Moreover, we answer a question: even if it is efficient, does it work well on applications?
To demonstrate the applicability, we conducted experiments using different models and datasets.


\paragraph{Setup}
First, we used the Stanford Sentiment TreeBank (SST-2)~\cite{socher2013sst} binary sentiment classification task.
Five thousand instances were sampled from the training set, and 872 instances in the development set were used.
We trained BERT-base classifiers~\cite{hugging2019transformer} with the adapter modules~\cite{houlsby19adapter}, which froze the pre-trained BERT parameters but newly trained branch networks in addition to the output layers.
We applied the turn-over dropout on the adapter modules and output layers.

In addition, we used the CIFAR-10~\cite{krizhevsky09cifar10} 10-class image classification task, with the 50,000 training instances and 10,000 validation instances.
We trained the VGGNet19 classifier~\cite{simonyan15vgg} with the turn-over dropout.

Models were trained with the cross-entropy loss.
Further details of the setup are shown in Appendix~\ref{sec:appendix_experimental_details}.

\subsection{Side Effect on Model Performance}\label{sec:experiment_side_effect}

Note that turn-over dropout is not for improving the accuracy of models.
It gives the models the method of efficiently estimating the influence of each training instance.
A possible side effect is a deterioration of accuracy due to introducing instance-specific dropout with $p=0.5$\footnote{Dropout with $p=0.5$ is often used in various neural networks, especially on linear layers of them, and improves the accuracy. However, dropout on all layers could damage. It is also unclear how dropout with ``static'' masks effect because the idea is novel.}.
Thus, we first explored the change of classification accuracy when using the turn-over dropout.

For BERT with the adapter modules on SST-2, if we use a small dataset (N=5,000), the accuracy slightly decreased from the baseline model, from 90.0\% to 88.3\%.
If we use a larger dataset (N=20,000), the change is negligible; 90.5\% and 90.2\%.
Thus, in a case with large datasets, where we typically want to use turn-over dropout for efficiency, applying the turn-over dropout does not decrease the validation accuracy compared with the baseline.
However, when we use turn-over dropout on all layers of BERT \emph{without} the adapter modules using makes training unstable~\footnote{
The instability might be due to the critical interruption of information caused by the high dropout rate.
Therefore, a possible remedy is performing turn-over dropout only in the backward pass, or only on the difference from the pre-trained BERT, which can be seen as turn-over dropout using mixout~\cite{lee2020mixout}.
}.
Furthermore, the same is true for VGGNet on CIFAR-10.
Instead, we first applied the turn-over dropout only for all layers after the 11th layer, although this means early layers can learn all instances in the training dataset and make the turn-over dropout leaky\footnote{
\citet{asano2020a} demonstrated that early layers of CNN contained limited information about the statistics of images, and such low-level statistics can be learned even through a single image.
Based on the finding, we assumed that early layers did not fit each instance so much, and the effect of leakage was small.
}.
We found that VGGNet with turn-over dropout can overfit more than the baseline does; their accuracies are 86.2\% and 92.0\%, respectively.
If we add regularization using the original dropout, the accuracy is recovered to 91.3\%.
Thus, in some cases, we have to care about the decrease of model performances when using turn-over dropout.
While we experimented with the successful architectures only, exploring the side effect in various architectures and remedies is important future work.

\begin{figure}[t]
  \begin{center}
   \includegraphics[width=7.6cm]{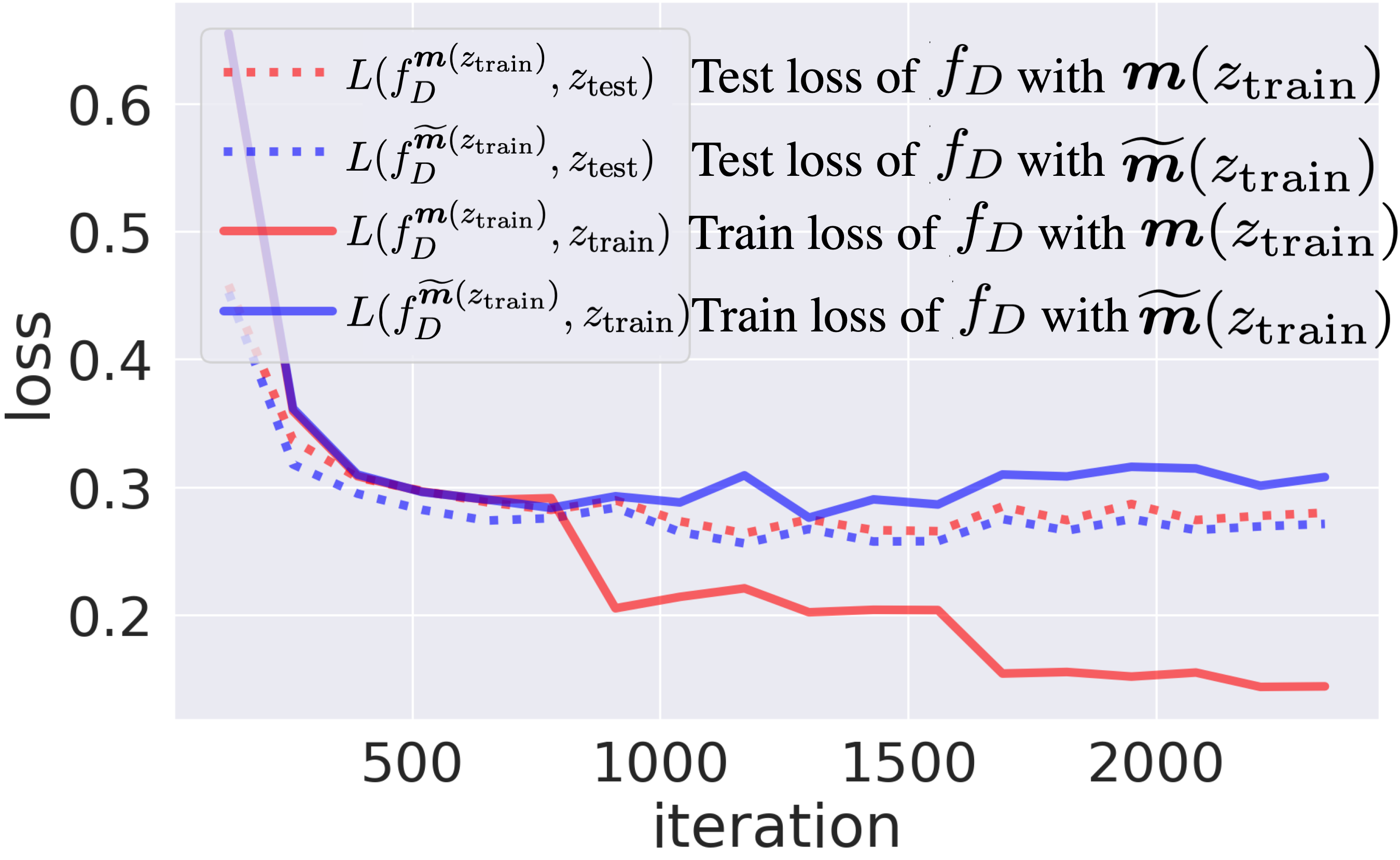}
    \end{center}
    \caption{Loss curves of BERT on SST-2.}
    \label{fig:sst2_loss_curves}
\end{figure}


\subsection{Learning Curves}\label{sec:learning_curves}

We first observed an interesting property of the turn-over dropout from the loss curves during training, as shown in Figure~\ref{fig:sst2_loss_curves}.
The solid red line of training loss using $\boldsymbol{m}(z_{\mathrm{train}})$, $L(f_{D}^{\boldsymbol{m}(z_{\mathrm{train}})}, z_{\mathrm{train}})$, showed a typical tendency of training loss.
However, the solid blue line of training loss using $\widetilde{\boldsymbol{m}}(z_{\mathrm{train}})$, $L(f_{D}^{\widetilde{\boldsymbol{m}}(z_{\mathrm{train}})}, z_{\mathrm{train}})$, indicated loss values close to the test losses (in dotted lines), without overfitting.
This fact agrees with the idea behind the turn-over dropout; the sub-network $f^{\widetilde{\boldsymbol{m}}(z_{\mathrm{train}})}$ using the flipped mask does not learn each training instance $z_{\mathrm{train}}$.

\begin{figure}[t]
  \begin{center}
  \includegraphics[width=7.6cm]{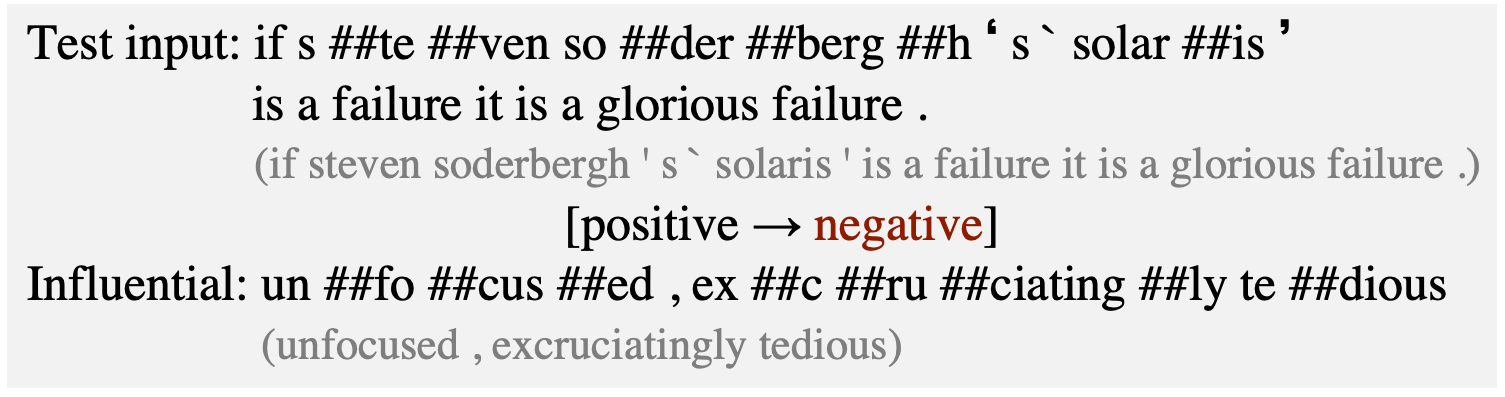}
    \caption{A misclassified text in the test set and the text with the highest influence with the error label in the training set for BERT on SST-2.}
    \label{fig:sst2_interpretation}
  \end{center}
\end{figure}

\begin{figure}[t]
  \begin{center}
  \includegraphics[width=7.6cm]{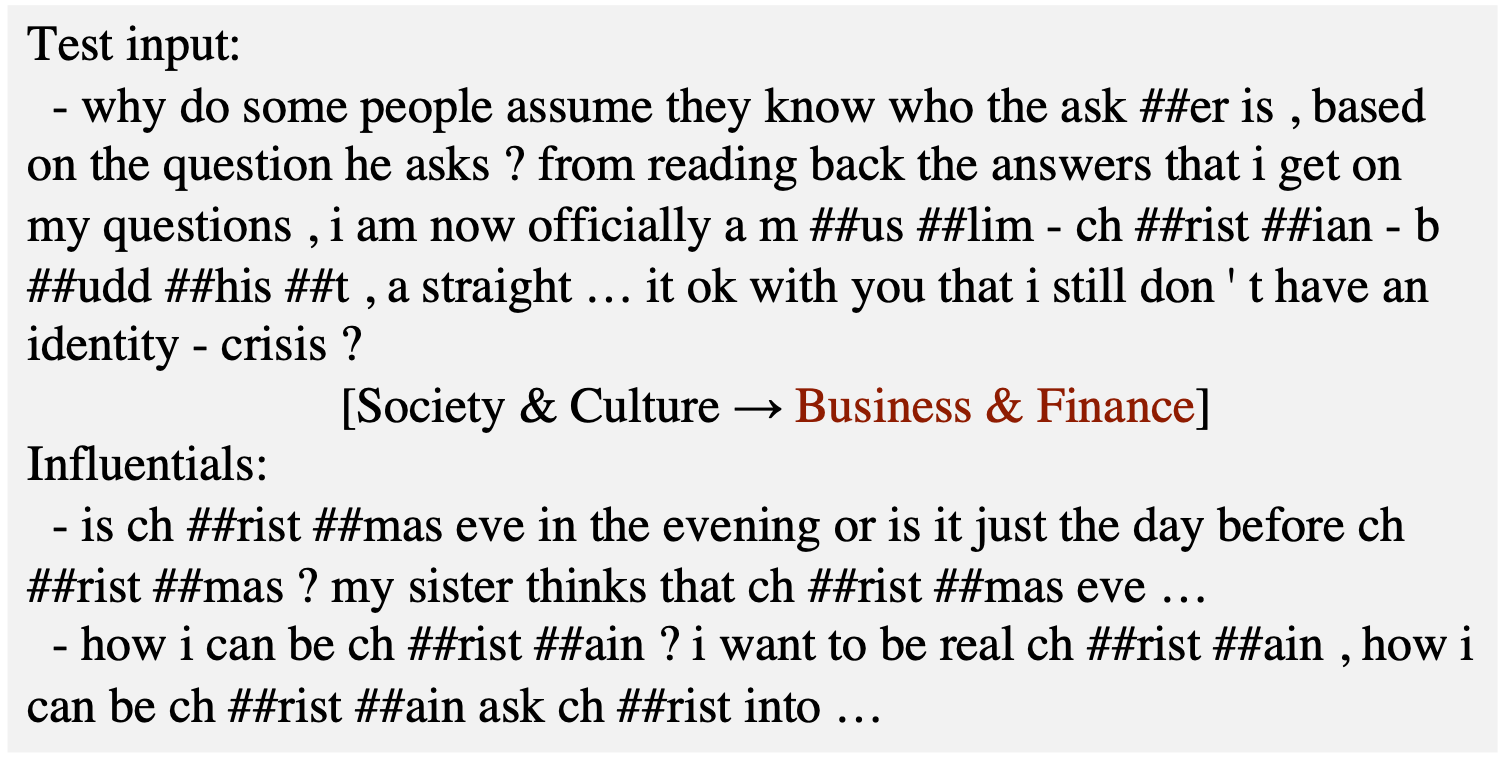}
    \caption{A misclassified text in the test set and the texts with the highest influence with the error label in the training set for BERT on Yahoo Answers.}
    \label{fig:yahoo_interpretation}
  \end{center}
\end{figure}

\begin{figure}[t]
  \begin{center}
  \includegraphics[width=7.6cm]{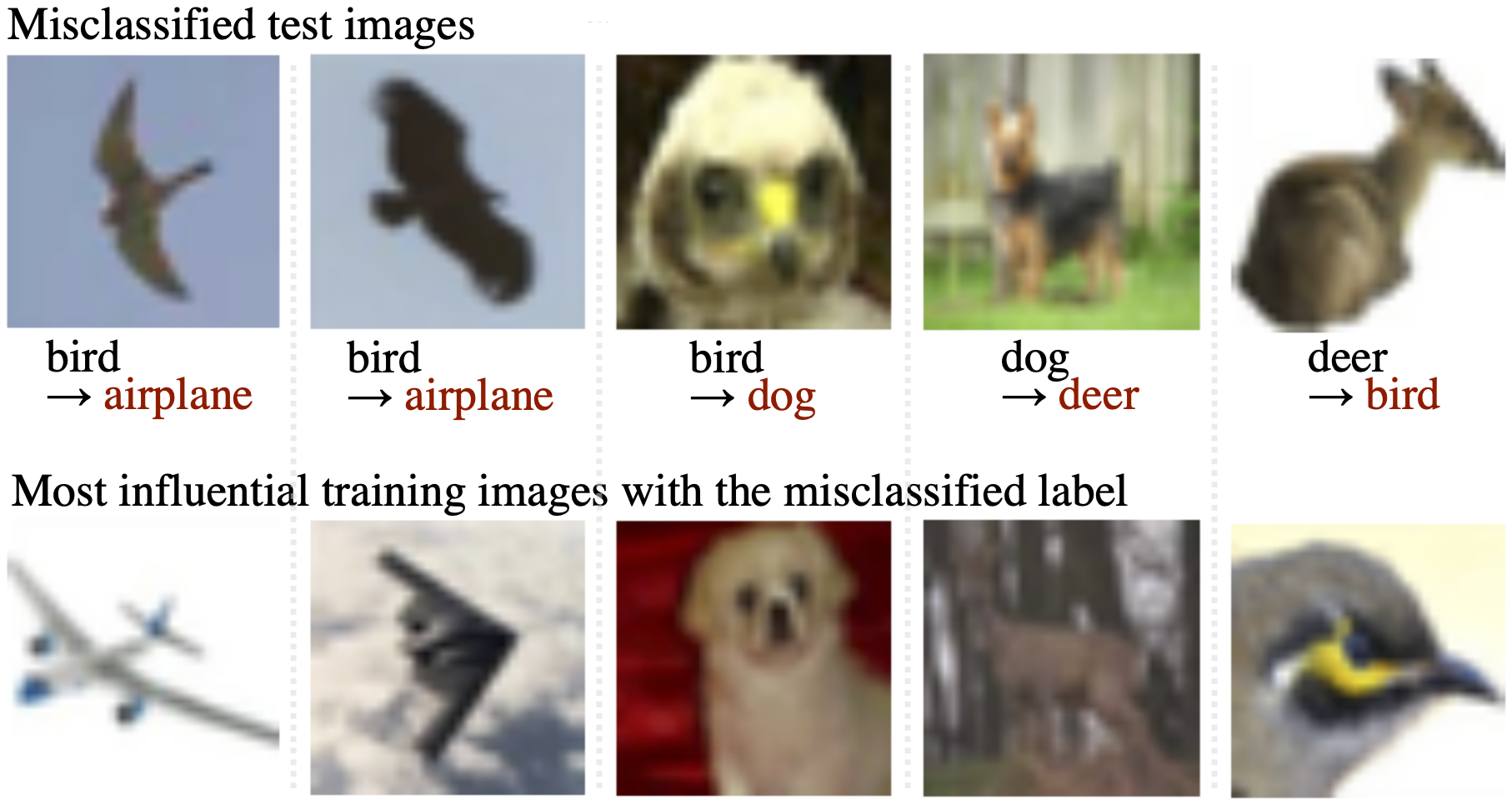}
    \caption{Misclassified images in the validation set (upper row) and images with the highest influence in the training set (lower row) for VGGNet on CIFAR-10.}
    \label{fig:cifar10_interpretation}
  \end{center}
\end{figure}


\subsection{Interpretation of Error of Predictions}\label{sec:experiment_interpretation}

Neural network models are notorious for their black-box prediction, which harms the trust and usability~\cite{ribeiro16trust}.
The influence estimation can mitigate this problem by suggesting possible reasons for a wrong model prediction by identifying influential training instances.

To verify this benefit, we collected the misclassified instances of the validation or test set and searched for the training instances that most influenced the wrong predictions.
Figure~\ref{fig:sst2_interpretation} indicates a text example from the results.
Rare words of named entities were divided into many subwords~\cite{shuster12wordpiece,sennrich-etal-2016-neural,wu16gnmt} and requiring more complex processing.
A guess is that BERT might fail to understand the input due to the cluttered subwords and predict a wrong label, which depended on a training instance similarly with many subwords.
Additionally, we conducted the same experiment on Yahoo Answers 10-label question classification dataset~\cite{zhang2015yahoo}\footnote{We used 5,000 training instances as well as SST-2.}, which is more complex than sentiment analysis.
Figure~\ref{fig:yahoo_interpretation} shows the results on Yahoo Answers.
The misclassified text shares the phrase ``ch \#\#rist'' with the two influential instances.
Such a low-level cue is not critical in the test.
However, it seemed that the model focused on the phrase and predicted the label of training instances containing the phrase.

In addition, more intuitively, image results are shown in Figure~\ref{fig:cifar10_interpretation}.
The two leftmost instances with the ``bird'' label were wrongly predicted as ``airplane.''
The training instances of airplane with the highest influence on the error predictions are shown in the row below.
The corresponding images had similar visual features, such as shape, layout, or color, which probably led to the wrong predictions.


\subsection{Data Cleansing}\label{sec:experiment_datacleasing}

\begin{table}[t]
\small
\begin{tabular}{lll}
\hline
                   & Accuracy (\%)               & Loss               \\ \hline
1\% Random Removal           & 76.8 $\pm$ 1.1          & 0.521 $\pm$ 0.030                \\
No Cleansing           & 77.0 $\pm$ 0.9          & 0.536 $\pm$ 0.063                \\ 
\textbf{1\% Cleansing} & \textbf{78.3 $\pm$ 0.2} & \textbf{0.484 $\pm$ 0.008}       \\ \hline
\end{tabular}
\caption{The results of data cleansing. Loss is the cross entropy loss. The averages and standard deviations from four difference runs are shown.}
\label{tab:data_cleansing_table}
\end{table}

Another possible application of the influence estimation is to eliminate harmful instances from the training dataset.
If the mean influence of a training instance on unseen instances is negative,
the instance can be harmful for generalization.
We experimented with data cleansing in a case of domain shift, where the training dataset is of SST-2 (movie review); however, the validation and test dataset are of the `electronics' subset in Multi-Domain Sentiment Dataset~\cite{blitzer-etal-2007-biographies} (Elec).
We split the Elec dataset into 200 instances for validation and 1,800 instances for the test.
Note that we do not use Elec dataset as a training dataset for studying only the effect of data cleansing.

We finetuned BERT models (with turn-over dropout) on SST-2 dataset and calculated the mean influences considering Elec's validation set.
After that, we re-trained models without turn-over dropout on datasets that removed training instances with 1\% of the most negative influences.
Finally, the model performances on Elec's test dataset are compared, as shown in Table~\ref{tab:data_cleansing_table}.
The models trained on the cleansed datasets achieved better accuracy and lower loss than those trained on the original dataset.
This result demonstrated that our estimation of the influence could also be used for data cleansing.


\section{Conclusion}

This paper proposed a method that required a low computational cost for estimating the influence of a training instance.
The method alters dropout with instance-specific masks and, for estimation,  uses sub-networks that are not trained with each instance.
The experiments demonstrated that this method could be applied even for complex models.

\section*{Acknowledgments}

We appreciate the helpful comments from the anonymous reviewers.
We thank Sho Takase, Hiroshi Noji, Hitomi Yanaka, Koki Washio, Saku Sugawara, Benjamin Heinzerling, and Kazuaki Hanawa for their constructive comments.
This work was supported by JSPS KAKENHI Grant Number JP19H04162.

\bibliography{anthology,emnlp2020}
\bibliographystyle{acl_natbib}

\clearpage
\appendix


\section{Self-influence by Language Model}\label{sec:experiment_language_modeling}

\begin{figure}[t]
  \begin{center}
   \includegraphics[width=7.6cm]{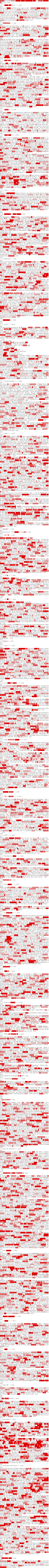}
  \end{center}
  \caption{Self-influence on WikiText-2 by a language model.}
  \label{fig:self_influence_lm}
\end{figure}

We also explored a new application of influence estimation for analyzing a language model and its training corpus.
This is the first work to use influence estimation for language modeling or other text generation tasks.
We trained a feed-forward neural language model~\cite{bengio2000lm}, which took as input fixed-length context words and predicted its next word; i.e., an instance is a pair of context words and its next word.
For analysis, we calculated the influence of each instance on a prediction of the instance itself, which we call \emph{self-influence}.
A concurrent work to ours \cite{feldman2020memorize} also propose the almost same value as an indicator of memorization.
They estimated the value by training thousands of models through using various subsets of the training dataset, and analyzed the behavior of the deep neural networks.
In language modeling, the higher self-influence of an instance is, the more easily a model would predict its next word if trained with it.
In other words, self-influence could indicate how effectively the instance can be memorized (or overfitted) and how difficult obtaining knowledge related to the instance from other instances is.
On language modeling, a typical case is when both the context and the next include low-frequency words.
In Figure~\ref{fig:self_influence_lm}, we show a snippet of WikiText-2 corpus~\cite{merity2016pointer} colored with the calculated self-influence values~\footnote{
For simplicity, the values are clipped to the range [70\% percentile, 95\% percentile], and calculated as average by ten models from different random seeds.
}.
A salient pattern is shown in the middle of the text, which lists the cast of a play; ``{\it The cast was : [eos] John Worthin ...}''.
Many last names have large self-influence values ({\it Worthing, Moncrieff, Vincent, Kinsey, Blacknell, Fairfax}, ...), because they are often infrequent and their contexts (i.e., first names) are also infrequent.
Such instances with large self-influence could be bad instances for generalization or lead to privacy issues~\cite{song2019auditdata,carlini2021extractingtraining} because overfitting or memorization can cause overestimation of the probability of specific words.
Interestingly, \citet{feldman2020memorize} empirically showed that training with examples of high self-influence were effective for test performance on image classification tasks.
The results could be different for language modeling because their image classification tasks and datasets intrinsically contain only less ambiguous labels (i.e., their `true' probability is almost one-hot) although language modeling does not satisfy the conditions at all and has an extreme long-tail property.
It will be intriguing to explore the analysis of self-influence values and feedback to improve language modeling training (e.g., with importance weighting).

The implementation is derived from \url{https://github.com/floydhub/word-language-model}.
The embedding and hidden layer size are 650, and dropout (p=0.3) and weight decay (0.00001) are applied for regularization.
The training epoch is 40, while the learning rate starts from 20 and is decayed (0.25) when validation perplexity is not improved.
The test perplexity after training is around 176~\footnote{Note that the performance of feed-forward neural network language models is typically much worse than RNN~\cite{merity2016pointer} or Transformer-based ones.}.

\section{Experimental Details}\label{sec:appendix_experimental_details}

In this section, we show the details of the experimental setting used in Section~\ref{sec:experiments}.

In this work, the BERT model, which is called \textit{bert-base-cased} in the Transformers library of \citet{hugging2019transformer}, was used.
The hyperparameters were selected based on the validation accuracy in preliminary experiments.
These hypterparameters were: learning rate = 5e-5 (from [2e-5, 5e-5]), batch size = 32, optimization epoch = 3 (from [3, 6, 10]), which were chosen by grid search based on the validation accuracy of SST-2.
For the models using the turn-over dropout, 10 epochs are used; while even training of 3 epochs worked and achieved the same accuracy, 10 epochs were a little more stable for estimation.

For the learning curves in Figure~\ref{sec:learning_curves}, the model is trained with 25,000 training instances for presenting the curves with less noise.

For the dataset of Multi-Domain Sentiment Dataset~\cite{blitzer-etal-2007-biographies} can be downloaded from \url{http://www.cs.jhu.edu/~mdredze/datasets/sentiment/}.
We extracted the `electronics' subset from their unprocessed dataset and tokenized the texts using Stanza~\cite{qi2020stanza} to align the input format as the SST-2 is already tokenized.

VGGNet19 was trained with the momentum SGD method with momentum = 0.9 and weight decay = 5e-4.
Moreover, a decaying learning rate by 0.1 was applied at the 150th and 225th epochs from the initial rate = 0.1, without the data augmentation of the horizontal flip.
The implementation was derived from \url{https://github.com/kuangliu/pytorch-cifar}.

\section{Runtime Compared with \citet{koh17understanding}}\label{sec:appendix_kohhan_comparison}

\citet{han2020explaining} reported that \citet{koh17understanding}'s method for BERT on the Multi-Genre NLI dataset~\cite{williams2018mnli} took 10 minutes to estimate the influences of 10,000 training instances on another single instance, using one NVIDIA GeForce RTX 2080 Ti GPU.
In our experiment, our proposed method took 35 seconds (i.e., 17 times faster) to estimate the same influences from the same dataset on the same GPU in our environment too.
While some accidental implementations may differ, both implementations of BERT are derived from \citet{hugging2019transformer}'s one.

In addition to the efficiency indicated by the big-O notation in Table~\ref{tab:computation_order_table}, our method can also process different training instances in a mini-batch efficiently at once because it only uses forward computations, unlike the others.
This is another advantage of our method in terms of efficiency.

\section{Hash-based Mask Composition}\label{sec:appendix_hashmask}

The volatile mask generation method (Section~\ref{sec:mask_construction}) solved storage and memory issues in our experiments.
However, memory issues (i.e., high space complexity) could occur even with the method, depending on implementations and dataset sizes.

For example, as a typical efficient implementation, mask generation for all instances in a minibatch should be performed with a few operations.
We can implement it with the two operations; at each layer during the forward computation, we (1) generate all instance-specific masks from a random seed and (2) extract a subset, which is required for the current mini-batch, by indexing.
In this case, the first step temporally requires a large memory space, which depends on the dataset size and the layer's dimension, while it volatilizes after the second step.
As a solution for mitigating memory usage, we can use \emph{hash-based mask composition}.
The basic idea is that we can generate different $N$ random masks from combinations of $K$ ($\ll N$) random masks.

We first generate a codebook composed of $K$ binary random masks, each of which is a $d$-dimensional dropout mask sampled from $\textrm{Bernoulli}(1 - p^{1/k})$.
We also prepare a hash function $H\colon Z \to \{1, ..., K\}^k$, which deterministically converts an instance $z\in Z$ to $k$ integers so that we can pick $k$ rows of the codebook.
Using the two components, given an instance, we can deterministically obtain $k$ primitive dropout masks, whose elements' ratio of 1 is $1 - p^{1/k}$.
Finally, performing cumulative product (or logical-AND) of the $k$ masks, we obtain a binary mask whose ratio of 1 is $p$ in expectation. After scaling it with the factor $\frac{1}{p}$, we can use it as a dropout mask.
This procedure can be implemented as fast batch processing using typical array operations only.
We can share the codebook with different layers in a network if using different hash functions so that different layers use different dropout masks.
Therefore, the space complexity of this algorithm is only $O(K\delta)$, where $\delta$ is the maximum dimension of a layer in a network.
And also, since the codebook can be used based on the volatile mask generation, it can avoid saving this codebook, i.e., reducing the required storage size to a minimum.

In summary, volatile mask generation reduces storage space and long-term memory space, and hash-based mask composition reduces both long and short-term memory spaces, while both require some computations for generation or composition.
The two techniques make instance-specific parameters applicable to large datasets.
These tricks are also used in \citet{takase2020word} for generating random word embeddings simultaneously.

\section{Self-influence of Training Instances}\label{sec:appendix_experiment_self_train}

\begin{figure}[t]
  \begin{center}
   \includegraphics[width=7.6cm]{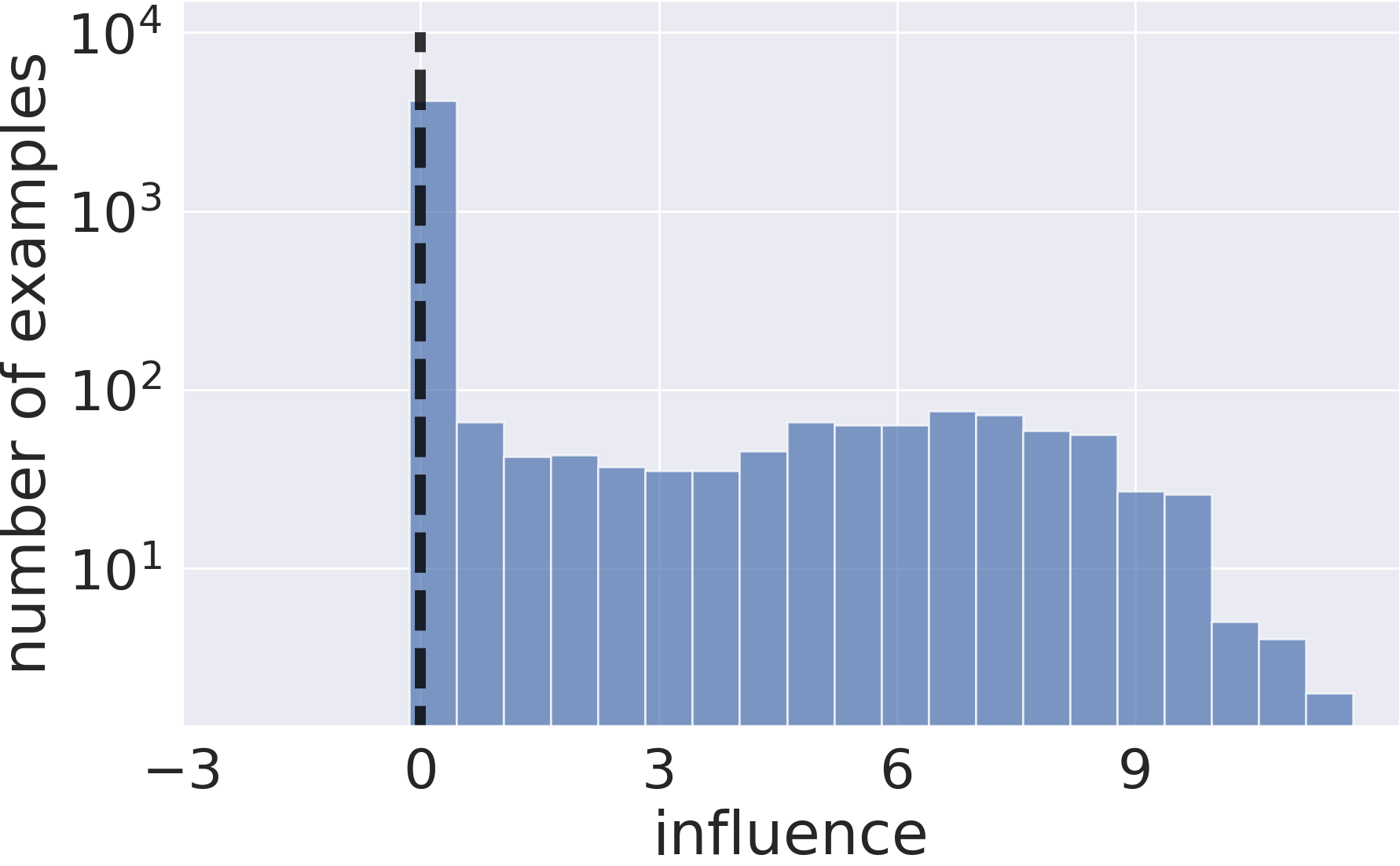}
  \end{center}
  \caption{Histogram of self-influence of BERT on SST-2.}
  \label{fig:histogram_sst2}
\end{figure}

We have conducted a preliminary experiment to analyze the estimated influences.
A common belief in supervised learning is that the model should achieve lower loss on the training instances.
For validating it, for BERT on SST-2, we estimated the influence of each training instance on a prediction of the instance itself (\emph{self-influence}) and presented histograms in Figure~\ref{fig:histogram_sst2}.
VGGNet also showed a similar distribution.
We can see that most of the instances have positive ($>0$) influence on themselves.
The results agree with the hypothesis that most of the training instances have positive influences on themselves.

\end{document}